\documentclass[a4paper]{article}

\usepackage{color}
\usepackage[usenames,dvipsnames,svgnames,table]{xcolor}
\usepackage{graphicx}
\usepackage{amsmath,amsthm,amssymb}
\usepackage[para]{threeparttable}
\usepackage{booktabs}
\usepackage{hyperref,url}

\usepackage{eulervm}
\usepackage{charter}
\usepackage{float}

\newcommand{\p}{\mbox{\,.}}

\newcommand{\B}{{\mathbb B}}
\newcommand{\cG}{{\mathcal G}}
\newcommand{\cS}{{\mathcal S}}

\newcommand{\N}{{\mathbb N}}

\newcommand{\I}{{\mathcal I}}

\newcommand{\hide}[1]{}
\newcommand{\tab}{\hspace*{5mm}}

\newcommand{\hs}{\hspace{0.1666em}} 

\floatstyle{ruled}
\newfloat{pseudo}{h}{lop}
\floatname{pseudo}{Algorithm}

\title{Large-scale network motif analysis using compression}

\author{Peter Bloem, Steven de Rooij}

\begin{document}

\maketitle

\begin{abstract}

 \noindent We introduce a new method for finding \emph{network motifs}: interesting or informative subgraph patterns in a network. Subgraphs are motifs when their frequency in the data is high compared to the expected frequency under a \emph{null model}. To compute this expectation, a full or approximate count of the occurrences of a motif is normally repeated on as many as 1000 random graphs sampled from the null model; a prohibitively expensive step. We use ideas from the Minimum Description Length (MDL) literature to define a new measure of motif relevance. With our method, samples from the null model are not required. Instead we compute the probability of the data under the null model and compare this to the probability under a specially designed alternative model. With this new relevance test, we can search for motifs by random sampling, rather than requiring an accurate count of all instances of a motif. This allows motif analysis to scale to networks with billions of links.
\end{abstract}


\section{Introduction}

Graphlets are small, induced subgraphs in a large network. \emph{Network motifs} \cite{milo2002network} are those graphlets that occur more frequently in the data than expected. To be able to conclude that such frequent subgraphs really represent meaningful aspects of the data, we must first show that they are not simply a product of chance. That is, a subgraph may simply be a frequent subgraph in \emph{any} random graph: a subgraph is only a \emph{motif} if its frequency is higher \emph{than expected}.

This expectation is defined in reference to a \emph{null model}: a probability distribution over graphs. We determine what the expected frequency of the subgraph is under the null model, and if the observed frequency is substantially higher than this expectation, the subgraph is a motif.

Unfortunately, there is usually no efficient way to compute the expected frequency of a subgraph under the null model. The most common approach generates a large number of random graphs from the null model and compares the frequencies of the subgraph in this sample to its frequency in the data \cite{milo2002network}. This means that any resources invested in extracting the motifs from the data must be invested again 1000 times to find out which subgraphs are motifs.

We introduce an alternative method that does not require such sampling from the null model. Instead, we use two probability distributions on graphs: the null model $p^\text{null}$, and a distribution $p^\text{motif}$ under which graphs with one or more frequent subgraphs have high probability. If a subgraph $M$ of a given graph $G$ allows us to show that $p^\text{motif}(G)$ is larger than $p^\text{null}(G)$, then $M$ is a motif. 

To design $p^\text{motif}$, we use the Minimum Description Length (MDL) Principle \cite{rissanen1978modeling,grunwald2007minimum}. We design a description method for graphs, a \emph{code}, which uses the frequent occurrence of a potential motif $M$ to create a compressed description of the graph. Our approach is analogous to compressing text by giving a frequent word a brief codeword: we describe $M$ once and refer back to this description wherever it occurs. 

By a commonly used correspondence between codes and probability distributions, we derive $p^\text{motif}$ from this code, a distribution that assigns high probability to graphs containing motifs.

Our approach speeds up motif analysis in two ways. First, we only need to compute $p^\text{null}(G)$ and $p^\text{motif}(G)$ instead of counting subgraphs in many random graphs. Second, it removes the need for accurate subgraph counts. For a potential motif $M$, we only need to find enough occurrences in  $G$ to achieve the required level of compression; we never need an exact count of the occurrences of $M$ in $G$. We simply sample random subgraphs until we find subgrahs with enough occurrences to yield a positive compression.

We show the following:
\begin{enumerate} 
\item Our method can be used to analyze graphs with millions of links in minutes. We can analyze graphs with billions of links in under 9 hours on a single compute node.
\item Our method can retrieve motifs that have been injected into random data, even at low quantities.
\item In real data, the motifs produced by our method are as informative in characterizing the graph as those returned by the traditional method. 
\end{enumerate}

%

Our exposition in this paper is relatively concise. We refer the reader to \cite{bloem2018tutorial} for a brief,  intuitive tutorial on using MDL for graph pattern analysis. All software is available open-source. \footnotemark 

\footnotetext{See \url{https://github.com/pbloem/motive} and \url{https://github.com/pbloem/motive-cls}.} 

\subsection{Related Work} 
Many different algorithms, techniques and tools have been proposed for the detection of motifs, all based on a common framework, consisting of three basic steps:

\begin{enumerate}
  \item Obtain a count $f_M$ of the frequency of subgraph $M$ in $G$. \label{step1}
  \item Obtain or approximate the probability distribution over the number of instances $F_M$ given that $G$ came from a particular null model $p^\text{null}$.  \label{step2}
  \item If $p^\text{null}(F_M \leq f_M) \leq 0.05$, consider $M$ a motif. \label{step3}
\end{enumerate} 

\noindent This was the approach proposed in \cite{milo2002network}, where the phrase \emph{network motif} was coined. One problem with this method is that it is very expensive to perform naively. Step \ref{step1} requires a full \emph{graph census}, and since the probability in step \ref{step3} cannot usually be computed analytically, we are required to perform the census again on thousands of graphs sampled from the null model in order to approximate it. 

Most subsequent approaches have attempted to improve efficiency by focusing on step~\ref{step1}: either by designing algorithms to get exact counts more efficiently \cite{koskas2011nemo,li2012netmode,khakabimamaghani2013quatexelero,meira2014acc}, or by approximating the exact count. The most extreme example of the latter is \cite{kashtan2004efficient}, which simply counts randomly sampled subgraphs. The complexity of this algorithm is \emph{independent of the size of the data}, suggesting an exceptionally scalable approach to motif detection. Unfortunately, while the resulting ranking of motifs by frequency is usually accurate, the estimate of their total frequency is not \cite{wernicke2005faster}, which makes it difficult to build on this approach in steps \ref{step2} and \ref{step3}. Other algorithms provide more accurate and unbiased estimates \cite{wernicke2005faster,ribeiro2010g,bhuiyan2012guise,jha2015path,slota2014complex,paredes2015rand}, but they do not maintain the scalability of the sampling approach.

We take an alternative route: instead of improving the sampling, we change the measure of motif relevance: we define a new hypothesis test as an alternative to steps \ref{step2} and \ref{step3}, which does not require an accurate estimate of the number of instances of the motif. All that is required is a set of some instances; as many as can be found with the resources available. This means that the highly scalable sampling approach from \cite{kashtan2004efficient} can be maintained.

The idea that compression can be used as a heuristic for subgraph discovery was also used in the SUBDUE algorithm \cite{cook1994substructure}. Our approach uses a more refined compression method and we connect it explicitly to the framework of motif analysis . We also exploit the possibility that the MDL approach offers, for a very scalable sampling algorithm, to replace the more restrictive beamsearch used in SUBDUE. 


%
The literature behind graph pattern mining, graphlets and network motifs seems to have developed largely in parallel, independently working towards  different goals. Graph pattern mining tends to focus on datasets consisting of many small graphs, rather than one large graph. As noted in \cite{aggarwal2014frequent}:
\begin{quote}
Defining the support of a subgraph in a set of graphs is straightforward, which is the number of graphs in the database that contain the subgraph. However, it is much more difficult to find an appropriate support definition in a single large graph  [\ldots].
\end{quote}
\noindent There are some efforts in the pattern mining literature to define new support measures, and other ways of efficiently counting frequent subgraphs. For the purposes of this paper, we will distinguish motif analysis from the broader field of pattern mining as follows: \emph{motif analysis refers to methods which aim to extract subgraphs characteristic for a single large graph, and which use the framework of hypothesis testing as a heuristic for this purpose}. Note that it is much harder to evaluate whether a method succesfully returns \emph{characteristic} subgraphs than it is to evaluate whether it returns \emph{frequent} subgraphs. We provide one way to operationalise this definition in Section~\ref{section:classification}.
For a good overview of recent work in graph pattern mining, we refer the reader to \cite[Chapter~13]{aggarwal2014frequent}. 
%
%
%
%

\subsection{preliminaries}

MDL is built on a very precise correspondence between optimizing for probability (learning) and optimizing for description length (compression). We will detail the basic principle below. For more information, see \cite{grunwald2007minimum,bloem2018tutorial}.
 
Let $\B$ be the set of all finite-length binary strings. We use $|b|$ to represent the length of $b \in \B$. Let $\log(x) = \log_2(x)$. A \emph{code} for a set of objects $\mathcal X$ is an injective function $f: {\mathcal X} \to \B$, mapping objects to binary code words. All codes in this paper are \emph{prefix-free}: no codeword is the prefix of another. We will denote a \emph{codelength function} with the letter $L$, ie. $L(x) = |f(x)|$. It is common practice to compute $L$ directly, without explicitly computing codewords and to refer to $L$ itself as a code.

The correspondence mentioned above follows from the \emph{Kraft inequality}: for any probability distribution $p$ on $\mathcal X$, there exists a prefix-free code $L$ such that for all $x \in \mathcal X$: $- \log p(x) \leq L(x) < -\log p(x) + 1$. Inversely, for every prefix-free code $L$ for $\mathcal X$, there exists a probability distribution $p$ such that for all $x \in \mathcal X$: $p(x) = 2^{-L(x)}$ \cite[Section~3.2.1]{grunwald2007minimum}, \cite[Theorem~5.2.1]{cover2006elements}. To explain the intuition, note we can transform a code $L$ into a sampling algorithm for $p$ by feeding the decoding function random bits until it produces an output. For the reverse, arithmetic coding \cite{rissanen1979arithmetic} can be used. 

As explained in \cite[page 96]{grunwald2007minimum}, the fact that $-\log p^*(x)$ is real-valued and $L^*(x)$ is integer-valued can be safely ignored and we may \emph{identify} codes with probability distributions, allowing codes to take non-integer values. 

 In some cases, we allow codes with multiple codewords for a single object, optionally indicating the choice for a particular codeword by a parameter $a$ as $L(x; a)$.

We focus on simple graphs: graphs containing no multiple links and no self-links. We consider both directed and undirected graphs. 

\section{MDL Motif analysis}

\label{section:motif-code}

The principle behind our method is simple: given two distributions $p^\text{null}$ and $p^\text{alt}$ we can show that for any data $G$
\begin{equation}
p^\text{null}\left[-\log p^\text{null}(G) + \log p^\text{alt}(G) \geq k\right ] \leq 2^{-k} \p \label{eq:inequality}
\end{equation}
The significance follows from the central principle of MDL: that any probability distribution $p^*(x)$ can be translated to a prefix-free code\footnotemark~which assigns $x$ a codelength of $L^*(x) = -\log(x)$ bits.

This means that (\ref{eq:inequality}), known as the \emph{no-hypercompression inequality} \cite[p103]{grunwald2007minimum}, can be interpreted as comparing codelengths: if we compress our data by a code corresponding to a chosen null model, and by \emph{any other code}, the probability that the alternative compresses better is exponentially small, as a function of the number of bits gained. In other words, if we see a compression gain of $k$ bits, we can reject the null model with a confidence of $1- 2^{-k}$.

To test whether a particular subgraph $M$ is a motif we proceed as follows. First, we compute the code-length $- \log p^\text{null}(G)$ of the data under a chosen null model (see section \ref{section:null models}). We then compress the graph using an ad-hoc motif code $p^\text{motif}$. If the latter compresses better than the former by $k$ bits, we may reject the null model.

In our experiments, we  will use not just a single null-model, but a lower bound $B^\text{null}$ on the codelength for all models in a particular set of models \cite{bloem2018tutorial}. If we achieve sufficient compression to beat the bound, we can reject all models in the set. This allows us to avoid any ad-hoc choices in the code for the null model, such as the code used for its parameters. For a more extensive explanation of this method, and the various subtleties in its use we refer the reader to \cite{bloem2018tutorial}.

\subsection{Motif Code}

We will now define the code used to compress graphs using a particular motif. We will use a given motif, and a list of its occurrences in the data to try to find an efficient description of the data. If this description allows us to reject a null model, we consider the motif interesting.\footnotemark

\footnotetext{Note that we have not ``found evidence'' for the motif as a pattern in any sense. We only use the hypothesis test as a \emph{heuristic}, as is the case in all motif analysis.}

Let $S = \langle S_1, \ldots, S_r \rangle$ be a sequence of nodes from $V_G$. Let $G[S]$ refer to the resulting induced subgraph of $G$.

\begin{figure}[tb]
  \includegraphics[width=\linewidth]{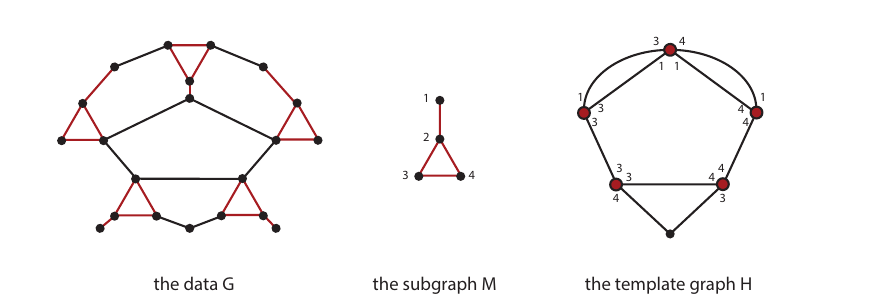}
  \caption{An illustration of the motif code. We store $M$ once, and remove its instances from $G$, replacing them with a single, special node. Storing $H$ and $M$, together with some``rewiring'' annotation, is enough to reconstruct $G$.}
   \label{figure:motif-code}
\end{figure}

Assume that we are given a graph $G$, a potential motif $M$, and a list $\I^\text{raw} = \langle I_1, \ldots, I_k\rangle$ of \emph{instances} of $M$ in $G$. That is, each sequence $I \in \I^\text{raw}$ consists  of nodes in $N_G$, such that the induced subgraph $G[I]$ is equal to $M$. Note that that $\I^\text{raw}$ need not contain all instances of $M$ in the data. Additionally, sequences in $\I^\text{raw}$ may overlap, i.e. two instances may share one or more nodes. We are also provided with a generic graph code $L^\text{base}(G)$ on the simple graphs. The basic principle behind our code is illustrated in Figure~\ref{figure:motif-code}.
%

\paragraph{Removing overlaps} The first thing we need is a subset $\I$ of $\I^\text{raw}$ such that the instances contained within it do not overlap: i.e. for each $I_a$ and $I_b$ in $\I$, we have $I_a \cap I_b = \emptyset$. 

An important factor for compression is the number of links an instance has to nodes outside the instance. We call this the \emph{exdegree}.\footnote{Unlike the in- and outdegree, the exdegree is not a property of a node, but of a subgraph.} We greedily remove all overlapping instances, always removing those with highest exdegree.

As part of the motif code, we will need to encode integers and sequences. We will use an integer model $p^\N$ and the Dirichlet-Multinomial model $p^\text{DM}$ for these purposes. Details are given in the appendix. 

\paragraph{The motif code} We can now define the full motif code. It stores the following elements. We use a prefix-free code for each, so we can simply concatenate the individual codewords to get a complete description of $G$.
\begin{description}
\item[subgraph] First, we store the subgraph $M$ using $L^\text{base}(M)$ bits.
\item[template] We then create the \emph{template graph} $H$ by removing the nodes of each instance $I \in \I$, except for the first, which becomes a specially marked node, called an \emph{instance node}. The internal links of $I$---those incident to two nodes both in $I$---are removed and links to a node outside of $I$ are rewired to the instance node.
\item[instance nodes] $L^\text{base}$ does not record which nodes of $H$ are instance nodes, so we must record this separately. Once we have recorded how many instance nodes there are, there are $n(H) \choose |\I|$ possible placements, so we can encode this information in $L^\N (|\I|) + \log {n(H) \choose |\I|}$ bits. 
\item[rewiring] For each side of a link in $H$ incident to an instance node, we need to know which node in the motif it originally connected to. Given some canonical order, we only need to encode the sequence $W$ of integers $w_i \in [1,\ldots, n(M)]$. 
\item[multiple edges] Since $L^\text{base}$ can only encode simple graphs, we remove all multiple edges form $H$ and encode them separately. We assume a canonical ordering over the links and record for each link incident to an instance node, how many copies of it were removed. This gives us a sequence $R$ of natural numbers $R_i \in [0, r_\text{max}]$ which we store by first recording the maximum value in $L^\N(\max(R))$ bits, and then recording $R$ with the DM model.
\item[insertions] Finally, while $H$ and $M$ give us enough information to recover a graph isomorphic to $G$, we cannot yet reconstruct where each node of a motif instance belongs in the node ordering of $G$. Note that the first node in the instance became the instance node, so we only need to record where to insert the rest of the nodes of the motif. This means that we perform $|\I| (n(M)-1)$ such insertions. Each insertion requires $\log (t+1)$ bits to describe, where $t$ is the size of the graph before the insertion. We require $\sum_{t=n(H)}^{n(G)-1} \log (t+1) = \log (n(G)!) - \log (n(H)!)$ bits to record the correct insertions.
\end{description}

\begin{pseudo}[tb]
\caption{The motif code $L^\text{motif}(G ; M, \I, L^\text{base})$. Note that the nodes of the graph are integers.}
\label{algorithm:motif-code}
{ 
\raggedright
Given:\\ 
\tab a graph $G$, a subgraph $M$,\\ 
\tab a list $\I$ of instances of $M$ in $G$, a code $L^\text{base}$ on the simple graphs.\\
~\\
$b_\text{subgraph} \leftarrow L^\text{base}(M)$\hfill \textbf{subgraph} \\
~\\
\emph{\# replace each instance with a single node} \\
$H \leftarrow \text{copy}(G)$, $W = [] $\hfill \textbf{template} \\
\textbf{for each } $I = \{ m_1, \ldots m_{n(M)}\}$ \textbf{in} $\I$:\\
\tab \emph{\# We use $m_1$ (the $m_1$-th node in $G$) as the instance node}\\
\tab \textbf{for each} link $l$ between a node $n_\text{out}$ not in $I$ and a node $m_j$ in $I$:\\
\tab \tab \textbf{if} $j \neq 1$: add a link between $n_\text{out}$ and $m_j$\\
\tab \tab $W$.append$(j)$\\
\tab remove all nodes $m_i$ except $m_1$, and all incident links\\
$b_\text{rewiring} \leftarrow  L^\text{DirM}_{|W|, n(M)}(W)\hfill\textbf{rewiring}$\\
~\\
\emph{\# remove multiple edges from $H$  and record the duplicates in $R$}\\
$R, H' \leftarrow \text{simple}(H)$ \\
$b_\text{template} \leftarrow L^\text{base}(H')$\\ 
$b_\text{multi-edges} \leftarrow L^\N(\max(R)) + L^\text{DirM}_{|R|, \max(R)}(R)$ \hfill \textbf{multiple edges}\\
~\\
$b_\text{instances} \leftarrow L^\N (|\I|) \log {n(H) \choose |\I|}$ \hfill \textbf{instance nodes}\\
$b_\text{insertions} \leftarrow \log (n(G))! - \log (n(H))!$ \hfill \textbf{insertions}\\    
~  \\
\textbf{return} $b_\text{subgraph} + b_\text{template} + b_\text{rewiring} + b_\text{multi-edges} + b_\text{instances} + b_\text{insertions}$\\
}
\end{pseudo} 
\paragraph{Pruning the list of instances} Since our code accepts any list of motif instances, we are free to take the list $\I$ and remove instances before passing it to the motif code, effectively discounting instances of the motif. This can often improve compression. We sort  $\I$ by exdegree and search for the value $c$ for which compressing the graph with only the first $c$ elements of $\I$ gives the lowest codelength.

The codelength $L^\text{motif}$ as a function of $c$ is roughly unimodal, so we use a \emph{Fibonacci search} \cite{kiefer1953sequential} to find a good value of $c$ while reducing the number of times we have to compute the full codelength.

%

\paragraph{Finding candidate motifs and their instances}

We search for motifs and their instances by sampling, based on the method described by \cite{kashtan2004efficient}. Since we do not require accurate frequency estimates we simplify the algorithm: start with a set $N'$ containing a single random node drawn uniformly. Add to $N'$ a random neighbour of a random member of $N'$, and repeat until $N'$ has the required size. Extract and return $G[N']$.

The size $n(M)$ of the subgraph is chosen before each sample from $U(n_\text{min}, n_\text{max})$. This distribution is biased towards small motifs: since there are fewer connected graphs for small sizes, small graphs are more likely to be sampled. The method still finds motifs with many nodes, so we opt for this simple, ad-hoc method.

We re-order the nodes of the extracted graph to a canonical ordering for its isomorphism class, using the Nauty algorithm \cite{mckay1981practical}. We maintain a map from each  subgraph in canonical form to a list of instances found for the subgraph. After sampling is completed, we end up with a set of potential motifs and a list of instances for each, to pass to the motif code.

\section{Null Models}

\label{section:null-models}
We will define three null models. For each, we first describe a parametrized model (which is not a code for  all graphs). We then use this to derive a bound so that we can reject a set of null models, and finally we describe how to turn the parametrized model into a complete model to store graphs within the motif code.  

Specifically, let $L^\text{name}_\theta(G)$ be a parametrized model with parameter $\theta$. Let $\hat\theta(G)$ be the value of $\theta$ that minimizes $L^\text{name}_\theta(G)$ (the maximum likelihood parameter). From this we derive a bound $B^\text{name}(G)$---usually using $B^\text{name}(G) = L^\text{name}_{\hat\theta(G)}(G)$---which we will use in place of the null model. Finally, we create the complete model by two-part coding: $L^\text{name}(G) = L^{\theta}(\hat\theta(G)) + L^\text{name}_{\hat\theta(G)}(G)$. 

\subsection{The Erd\H{o}s-Renyi Model}
\label{section:null models}

The Erd\H{o}s-Renyi (ER) model is probably the best known probability distribution on graphs \cite{renyi1959random,gilbert1959random}. It takes a number of nodes $n$ and a number of links $m$ as parameters, and assigns equal probability to all graphs with these attributes, and zero probability to all others. This gives us 
\begin{align*}
L^\text{ER}_{n, m}(G) = \log{n^2-n \choose m} & & L^\text{ER}_{n, m}(G) = \log{(n^2-n)/2 \choose m} 
\end{align*}
for directed and undirected graphs respectively. We use the bound $B^\text{ER}(G) = L^\text{ER}_{n(G), m(G)}(G)$.

For a complete code on simple graphs, we encode $n$ with $L^\N$. For $m$ we know that the value is at most $m_\text{max}=(n^2-n)/2$ in the undirected case, and at most $m_\text{max}=n^2-n$ in the directed case, and we can encode such a value in $\log (m_\text{max} + 1)$ bits ($+1$ because $0$ is also a possibility). This gives us:\\
$
L^\text{ER}(G) = L^\N(n(G)) + \log (m_\text{max} + 1) + L^\text{ER}_{\theta}(G)\;\text{with}\;\theta=(n(G),m(G))\p
$
 
\subsection{The Degree-Sequence Model}
\label{section:degree-sequence-model}

The most common null model in motif analysis is the \emph{degree-sequence model}, also known as the \emph{configuration model} \cite{newman2010networks}. 
For undirected graphs, we define the degree sequence of graph $G$ as the sequence $D(G)$ of length $n(G)$ such that $D_i$ is the number of links incident to node $i$ in $G$. For directed graphs, the degree sequence is a pair of such sequences $D(G) = (D^\text{in}, D^\text{out})$, such that $D^\text{in}_i$ is the number of incoming links of node $i$, and $D^\text{out}_i$ is the number of outgoing links.

\paragraph{The parametrized model $L^\textnormal{DS}_D(G)$} The degree-sequence model $L^\text{DS}_D(G)$ takes a degree sequence $D$ as a parameter and assigns equal probability to all graphs with that degree sequence. Assuming that $G$ matches the degree sequence, we have $L^\text{DS}_D(G) = \log |\cG_D|$ where $\cG_D$ is the set of simple graphs with degree sequence $D$. There is no known efficient way to compute this value for either directed or undirected graphs, but various estimation procedures exist. We use an importance sampling algorithm from  \cite{blitzstein2011sequential,charo2010efficient}.

\paragraph{The bound $B^\textnormal{DS}(G)$} We make the assumption that the degrees are sampled independently from a single distribution $p^\text{deg}(n)$ on the the natural numbers. This corresponds to a code $\sum_{D_i \in D}L^\text{deg}(D_i)$ on the entire degree sequence. Let $f(s, D)$ be the frequency of symbol $s$ in sequence $D$. It can be shown that $B^\text{deg}(D) = \sum_{D_i \in D} f(D_i, D)/|D|$ is a lower bound for any such code on the degree sequence. This gives us the bounds $B^\text{DS}(G) = B^\text{deg}(D(G))) + L^\text{DS}_{D(G)}(G)$ and $B^\text{DS}(G) = B^\text{deg}(D^\text{in}(G))) + B^\text{deg}(D^\text{out}(G))) + L^\text{DS}_{D(G)}(G)$.
%

\paragraph{The complete model $L^\textnormal{DS}(G)$} For the alternative model we need a complete code. First, we store $n(G)$ with $L^\N$. We then store the maximum degree and encode the degree sequence with the DM model. For undirected graphs we get: 
\begin{align*}
L^\text{DS}(G) = L^\N(n(G)) + L^\N(d) + L^\text{DirM}_{\theta}(D) + L^\text{DS}_{D(G)}(G) \\ \;\text{with}\; \theta = \left(n(G), max(D)\right)
\end{align*}
and for directed graphs
\begin{align*}
L^\text{DS}(G) = &L^\N(n(G)) \\ 
 + &L^\N(\max(D^\text{in})) + L^\text{DirM}_{\theta}(D^\text{in})\\
 + &L^\N(\max(D^\text{out})) + L^\text{DirM}_{\phi}(D^\text{out}) + L^\text{DS}_{D(G)}(G) \\
 \text{with}&\; \theta = \left(n(G), \max(D^\text{in})\right), \phi = \left(n(G), \max(D^\text{out})\right) 
\end{align*} 


\subsection{The Edgelist Model}

While estimating $|\cG_D|$ can be costly, we can compute an upper bound efficiently. Assume that we have a directed graph $G$ with $n$ nodes, $m$ links and a pair of degree sequences $D = (D^\text{in}, D^\text{out})$. To describe $G$, we might write down the links as a pair of sequences $(F, T)$ of nodes: with  $F_i$ the node from which link $i$ originates, and $T_i$ the node to which it points. Let $\cS_d$ be the set of all pairs of such sequences satisfying $D$. We have $ v$ possibilities for the first sequence, and $m \choose D_1^\text{out}, \ldots, D_n^\text{out}$ for the second. This gives us $|\cS_D| = {m \choose D_1^\text{in}, \ldots, D_n^\text{in}}{m \choose D_1^\text{out}, \ldots, D_n^\text{out}} = m!m! / \prod_{i=1}^n D^\text{in}_i ! D^\text{out}_i !$. We have $|\cS_D| > |\cG_D|$ for two reasons. First, many of the graphs represented by such a sequence pair contain multiple links and self-loops, which means they are not in $\cG_D$. Second, the link order is arbitrary: we can interchange any two different links, giving a different pair of sequences, representing the same graph. A graph with no multiple edges, is represented by $m!$ different sequence-pairs. 

To refine this upper bound, let $\cS'_D \subset \cS_D$ be the set of sequence pairs representing simple graphs. Since all links in such graphs are distinct, we have $|\cG_D| = |\cS'_D|/m!$. Since $|\cS'_D| \leq |\cS_D|$, we have
\[
|\cG_D| \leq \frac{m!}{\prod_{i=1}^n D^\text{in}_i ! D^\text{out}_i !} \p
\]


In the undirected case, we can imagine a single, long list of nodes of length $2m$. We construct a graph from this by connecting the node at index $i$ in this list to the node at index $m+i$ for all $i \in [1, m]$. In this list, node $a$ should occur $D_a$ times. We define $\cS_D$ as the set of all lists such that the resulting graph satisfies $D$. There are $(2m)! \choose D_1, \ldots, D_n$ such lists.
We now have an additional reason why $|\cS_D| > |\cG_D|$: each pair of nodes describing a link can be swapped around to give us the exact same graph. This gives us:
\[
|\cG_D| \leq |\cS'_D| / (2^m m!) = \frac{(2m)!}{2^m m! \prod_{i=1}^n D_i!} \p
\]


 This gives us the following parametrized code for directed graphs:
\begin{equation}
\label{eq:el-dir}
L^\text{EL}_D(G) = \log m! - \sum_{i=0}^n \log D_i^\text{in}! - \sum_{i=0}^n \log D_i^\text{out}!   
\end{equation}
where $(D^\text{in}, D^\text{out})$ are the degree sequences of $G$, and for the undirected case:
\begin{equation}
\label{eq:el-undir}
L^\text{EL}_D(G) = \log (2m)! - \log m! - m - \sum_{i=0}^n \log D_i! \p   
\end{equation}

For the bound and the complete model, we follow the same approach we used for the degree-sequence model.


\section{Experiments}

\label{section:experiments}


In all experiments, we report the \emph{log-factor}: 
\[
B^\text{null}(G) - L^\text{motif}(G; M, \I, L^\text{null}) \p
\]
This is a value in bits, indicating how much better the motif code compresses than the lower bound on the null model. If the log-factor is larger than 10 bits, we can interpret it as a successful hypothesis test at $\alpha=0.001$.\footnote{A negative log-factor means that we do not have sufficient evidence to reject the null model, but a different experiment might yet achieve a positive log-factor.}

\subsection{Recovering Motifs from Generated Data}

\label{section:recovering}

In our first experiment we will test some of the basic expected behaviors from our method: (1) a graph sampled randomly from a simple null model should contain many frequent subgraphs, but no motifs. (2) If a subgraph is manually inserted a number of times, we should be able to detect this graph as a motif. To test this, we sample a graph from a null model and inject $k$ instances of a specific motif, in a way that corresponds broadly to the motif code. Algorithm~\ref{algorithm:motif-code} provides the procedure in detail.

\begin{figure*}[tb]
  \includegraphics[width=\textwidth]{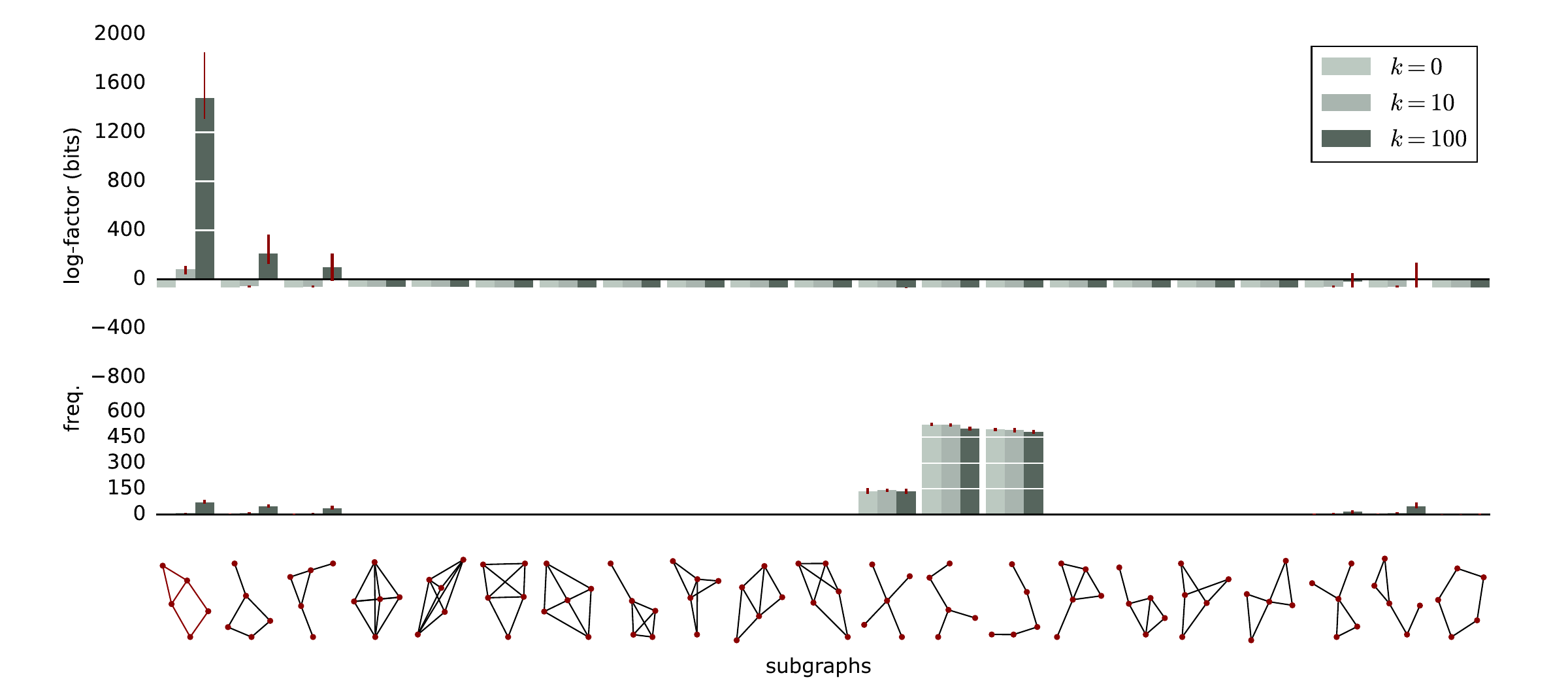}
  \caption{\small Motif analysis on generated data with $k$ inserted motifs for all 21 simple connected graphs with 5 nodes. The middle row shows the number of non-overlapping instances found by the sampling algorithm for each potential motif. The bars show the average value over 10 randomly sampled graphs, with the same subgraph (shown in red) injected each time. Error bars represent the \emph{range}.}
  \label{figure:plot-synthetic}
\end{figure*}


On this sampled graph, we run our motif analysis. We run the experiment multiple times, with $k = 0$, $k = 10$ and $k = 100$, using the same subgraph $M$ over all runs, but sampling a different $H$ each time. For each value of $k$, we repeat the experiment 10 times. Per run, we sample only  5000 subgraphs. 


Figure~\ref{figure:plot-synthetic} shows the results for the 21 possible connected simple graphs of size 5. This result shows that, on such generated data, the method behaves as expected in the following ways:
\begin{itemize}
	\item If no motifs are injected no subgraphs are motifs.
	\item Even for very low $k$, the correct motif is given a positive log-factor. Other subgraphs are shown to have very high frequencies, but a negative log factor. 
	\item If the number of motifs is high ($k = 100$), the resulting log-factor increases.
\end{itemize}

\noindent We can also see that once we insert 100 instances of the motif, two other subgraphs ``become motifs'': in both cases, these share a part of the inserted motif (a rectangle and a triangle). This effect is not unique to our method, but occurs in all motif analysis. 

The relative magnitude of the log factors provides a ranking within those subgraphs marked as motifs. In traditional motif methods, computing these relative magnitudes accurately requires very large samples of random graphs.

\subsection{Motifs from Real-World Data}
\label{section:various}

Next, we show how our approach operates on a selection of data sets across domains. Our main aim with this experiment is to show how the three null models influence the results. Specifically, to ascertain whether the edgelist model provides a reasonable approximation for the degree-sequence model. The data sets are described in the supplement.

\begin{figure}[tbh]
  \includegraphics[width=\linewidth]{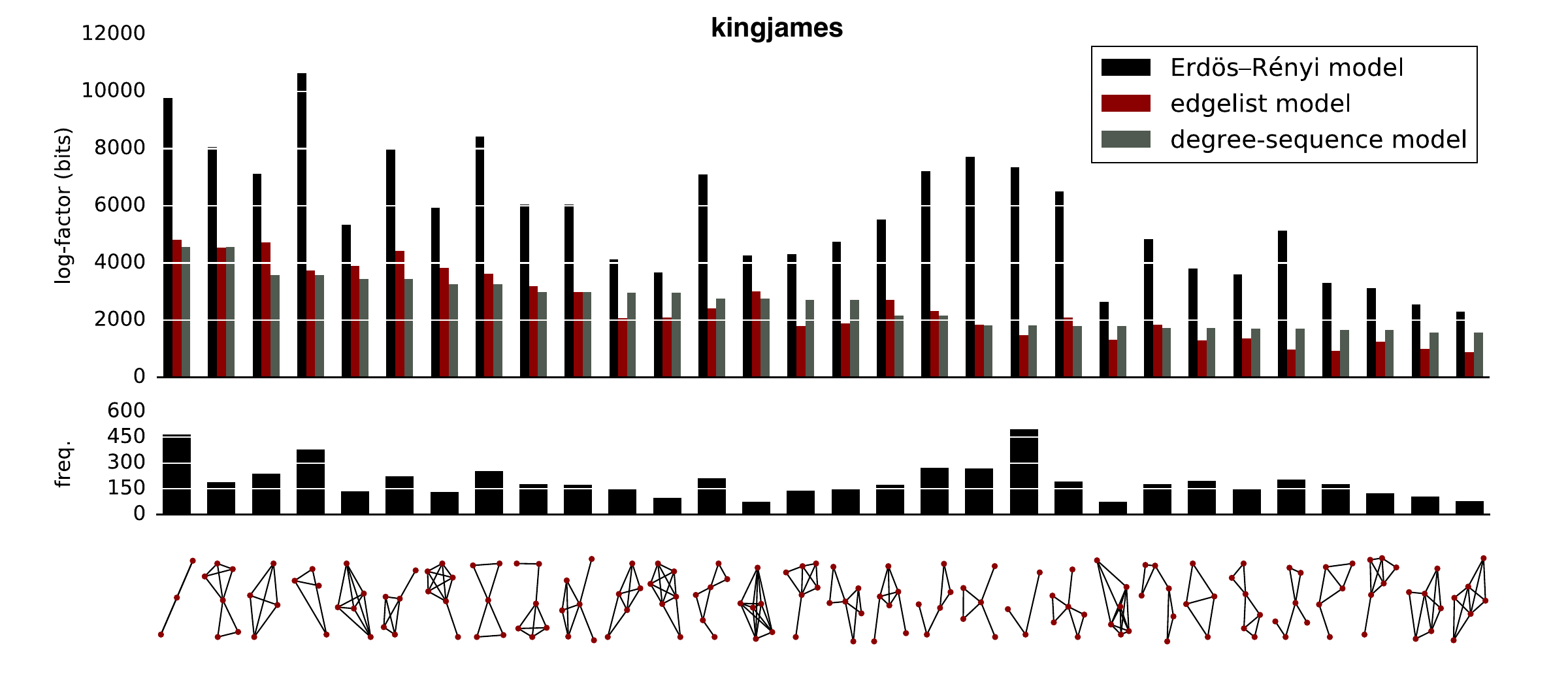}\\
  \includegraphics[width=\linewidth]{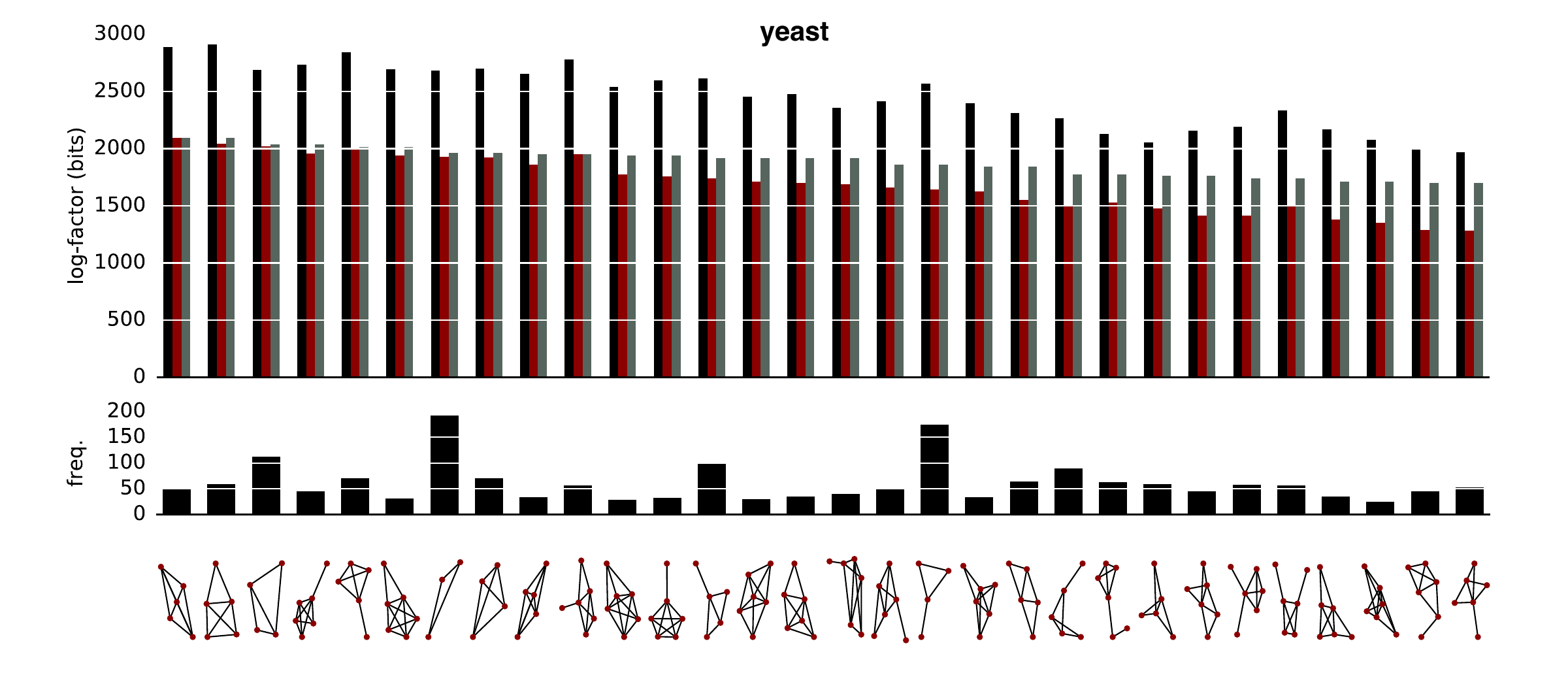}\\
  \caption{The results of the motif extraction on the 2 undirected networks.}
  \label{figure:plot-und}
\end{figure}
  
\begin{figure}[tbh]
  \includegraphics[width=\linewidth]{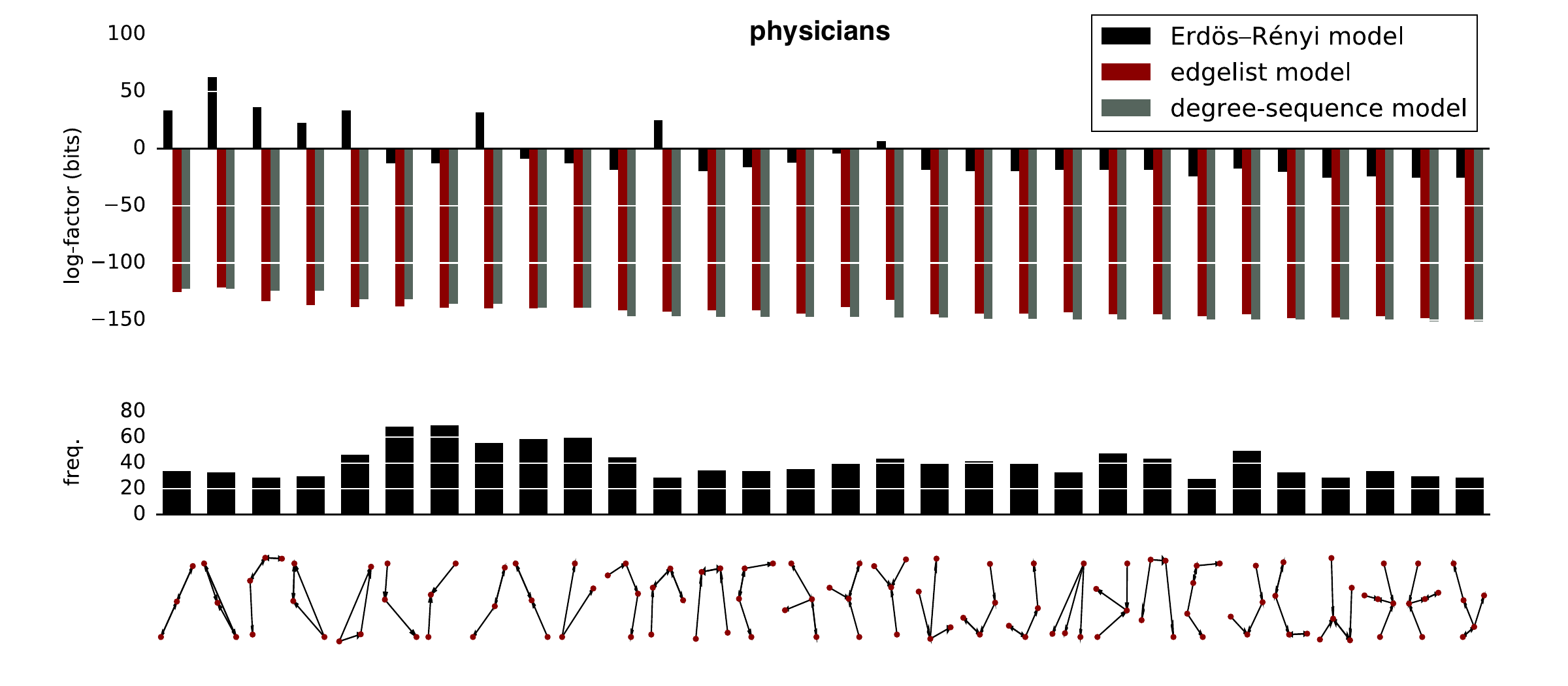}\\
  \includegraphics[width=\linewidth]{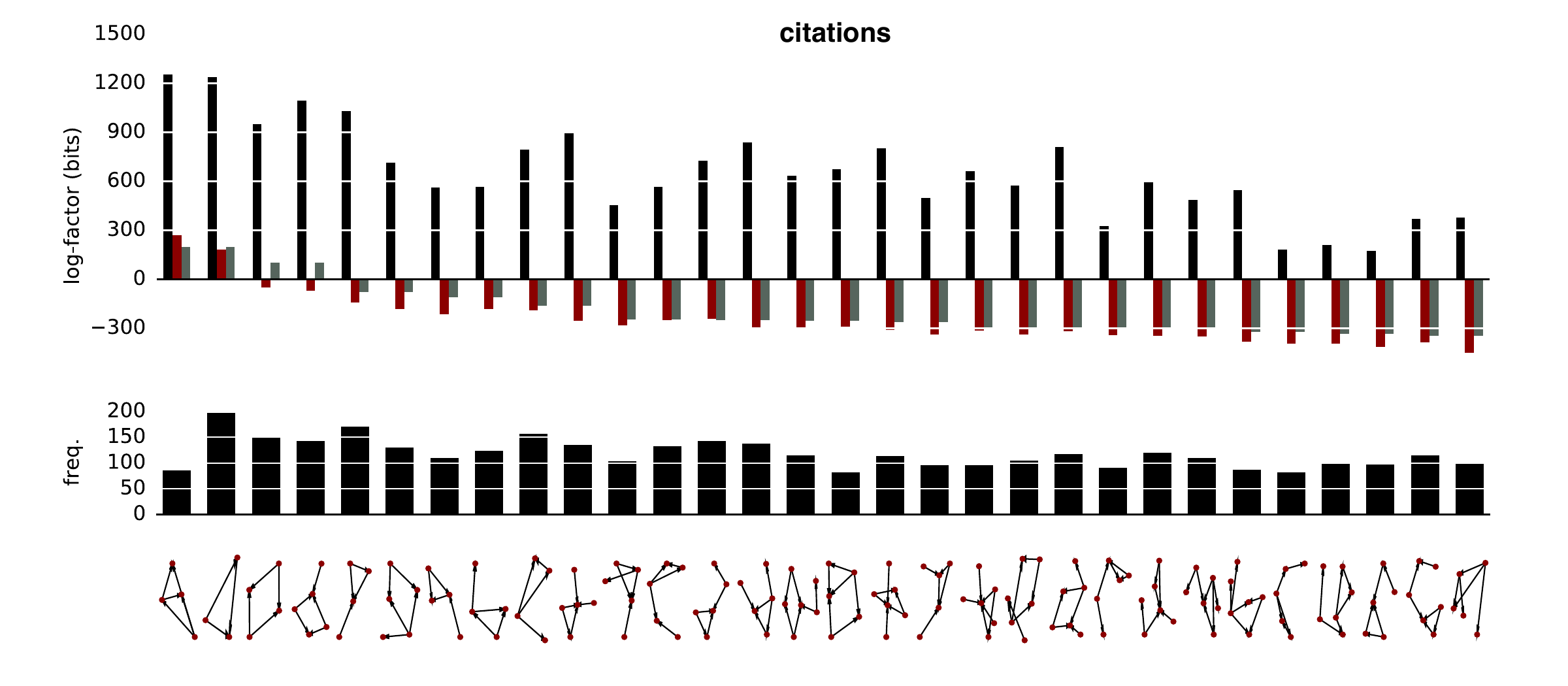}\\
  \caption{The results of the motif extraction on the 2 directed networks.}
  \label{figure:plot-dir}
\end{figure}


Our first observation is that for the physician data set, there are no motifs under the degree-sequence null model. This likely because the physicians network is too small: the use of a bound for the null model means that the alternative model requires a certain amount of data before the differences become significant. Note, however, that if we were to compare against a complete model (instead of the bound), a constant term would be added to all compression lengths under the null model. In other words, the ordering of the motifs by relevance would remain the same.

In both the kingjames and the yeast graphs, many motifs contain (near-)cliques. This suggests the data contains communities of highly interconnected nodes which the null model cannot explain.


For the experiments in this section, the maximum Java heap space was set to 2 Gigabytes. The computation of the log-factor of each motif was done in parallel, as was the sampling for the degree sequence model, with at most 16 threads runnning concurrently, taking advantage of the 8 available (physical) cores. 

These experiments took relatively long to run (ranging from 31 minutes to nearly 24 hours). The bottleneck is the computation of the degree-sequence model. If we eliminate that, as we do in Section~\ref{section:large}, we see that we can run the same analysis in minutes on graphs that are many orders of magnitude larger than these. Moreover, the plots show a reasonable degree of agreement between the EL model and the DS model, suggesting that the former might make an acceptable proxy. The next section tests whether the resulting motifs are still, in some sense, informative.

\subsection{Comparison with the traditional method}

\label{section:classification}

\begin{figure}[tbh]
{
  \includegraphics[width=\linewidth]{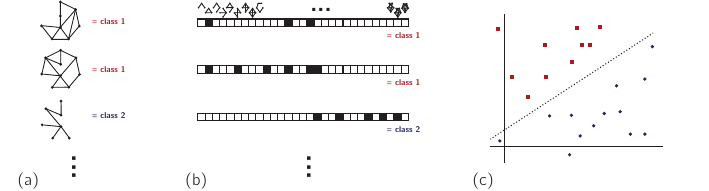}	
}

  \caption{A schematic illustration of the classification experiment. (a) We start with a classification task on simple, undirected graphs. (b) These graphs are reduced to 29-dimensional binary feature vectors. Each feature corresponds to an undirected, connected mini-graph of size 3, 4, or 5. (c) We apply a linear SVM to perform the classification.}
  \label{figure:experiment-explanation}
\end{figure}


The definition of what constitutes a motif is exceedingly vague: papers variously describe a motif as a ``functional unit'', a ``characteristic pattern'' or a ``statistically significant subgraph.'' To operationalise this to something that we can test empirically, we will define a network motif as a subgraph that is \emph{characteristic} for the full graph. That is, in some manner, the information that ``$M$ is a motif for $G$'' should \emph{characterize} $G$: it distinguishes $G$ from the graphs for which $M$ is not a motif, and that distinction should be meaningful in the domain of the data.

We operationalize ``making a meaningful distinction in the domain of the data'' as \emph{graph classification}. If motif judgments (as binary features) can be used to beat a majority-class baseline by a significant amount, we can be sure that they make a meaningful distinction in the domain of the data. 

We start with a set of undirected simple graphs, with associated classes. We then translate each graph into a binary feature vector using \emph{only the motif judgements} of the algorithm under evaluation. We test all connected subgraphs of size 3, 4, and 5, giving us 29 binary features. If a simple classifier (in our case a linear SVM) can classify the graphs purely on the basis of these 29 motif judgments, the algorithm has succeeded in characterizing the graph. 

For those algorithms that succeed, the classification accuracy can be used to measure relative performance, although we should not expect high performance in a task as challenging as graph classification from just 29 binary features.

This approach---quantifying unsupervised pattern extraction through \linebreak[4]classification---was also used in \cite{leeuwen2006compression}. 

Our main aim is to establish that the resulting classifier performs better than chance. Our secondary aim is to show that we do not perform much worse than the traditional method. 



For our purposes, we require classifications tasks in a narrow range of sizes: the graphs should be small enough that we can use the traditional method without approximation, but large enough that our method has enough data to confidently reject a null bound.

In order to tune the graph classification tasks to our needs, we adapt them from classification tasks on knowledge graphs \cite{ristoski2016collection}. In these, the data is a single labeled, directed multigraph, and the task is to predict classes for a specific subset of nodes (the \emph{instance nodes}). We translate the graph to an unlabeled simple graph by using the same nodes (ignoring their labels) and connecting them with a single undirected edge only if there are one or more directed edges between them in the original knowledge graph. 

This gives us a classification task on the nodes of a single, undirected simple graph. We turn this into a classification task on \emph{separate} graphs by extracting the 3-neighborhood around each instance node. To control the size of the extracted neighborhoods, we remove the $h$ nodes with the highest degrees from the data before extracting the neighborhoods. $h$ was chosen by trial-and-error, before seeing the classification performance, to achieve neighborhoods with around 1000--2000 nodes.

We now have a graph classification task from which we can create feature vectors as described above. For our method, we sample 100\hs000 subgraphs, with size 3, 4, 5 having equal probability and test the compression levels under the edgelist model. We judge a subgraph to be a motif if it beats the EL bound by more than $- \log \alpha$ bits with $\alpha = 0.05$. 

Many methods for motif analysis have been published, but most are approximations or more efficient counting algorithms. Therefore, a single algorithm based on exact counts can act as a baseline, representing most existing approaches: we perform exact subgraph counts on both the data and 1\hs000 samples from the DS model. The samples from the null model are taken using the Curveball algorithm \cite{strona2014fast,carstens2016curveball}. We estimated the mixing time to be around 10\hs000 steps, and set the run-in accordingly. The subgraph counts were performed using the ORCA method.\footnote{We created a Java implementation, available at \url{https://github.com/pbloem/orca}} We mark a subgraph as a motif if fewer than 5\% of the graphs generated from the DS model have more instances of the subgraph than the data does.\footnotemark~

\footnotetext{Note that the commonly used z-score method is seriously flawed, as discussed in \cite{picard2008assessing}, so we do not use it here.}

For performance reasons (we are at the limits of what the traditional method allows), we use only 100 randomly chosen instances from the classification task. On these 100 instances, we perform five-fold cross-validation. To achieve good estimates, we then repeat the complete experiment, from sampling instances to cross-validation, 10 times. The classifier is a linear SVM ($C=1$). For tasks with more than 2 classes, the one-against-one approach \cite{knerr1990single} is used.


\begin{figure}[tb]
{
  \begin{center}
  \includegraphics[width=\linewidth]{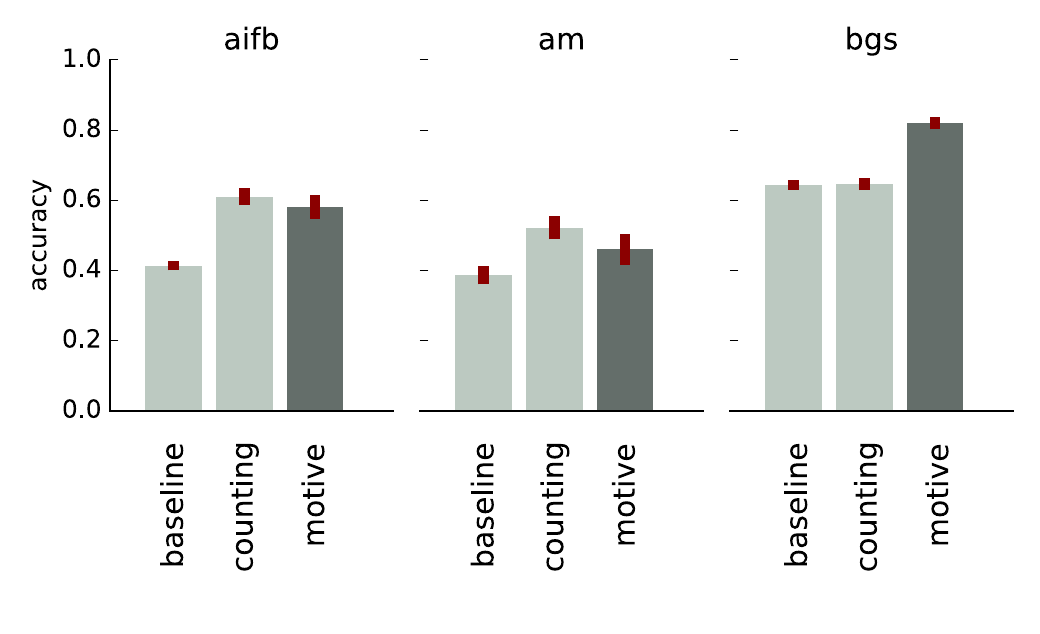}
	\begin{tabular}{ r | r r r r r r}
		data & \# nodes & \# links & $h$ & $K$ & $\overline{n}$ & $\overline{m}$ \\
		\hline
		AIFB & 8\hs275 & 17\hs911 & 10 & 4 & 1877.11 & 7\hs141.48 \\
		AM   & 1\hs495\hs566 & 2\hs393\hs604 & 3000 & 11 & 2506.07 & 4392.06 \\
		BGS  & 333\hs613 & 362\hs627 & 250 & 2 & 3097.47 & 4404.49 \\
		\hline
	\end{tabular}
	\end{center}	
}

  \caption{The results of the classification experiment for the traditional method (\emph{counting}) and ours (\emph{motive}). The bars show the mean accuracy of ten runs (with cross-validation within each run). Error bars show a 95\% confidence interval. The baseline is a majority-class classifier which ignores the features.  The table shows the size of the data, the average size and number of links ($\overline{n}$, $\overline{m}$) of the instance graphs, the number of classes $K$ and the number $h$ of hubs removed.}
  \label{figure:classification}
\end{figure}

The results are shown in Figure~\ref{figure:classification}. For one data set, our method is significantly better, for another, the traditional approach is significantly better, and for one, the difference is not significant. While the performance of neither method is stellar, the fact that both beat the baseline significantly, shows that at the very least, the motifs contain \emph{some} information about the class labels of the instance represented by the graph from which the motifs were taken.

\begin{table}
\centering\small 
\begin{threeparttable}[t]
\begin{tabular}{ l  r r r r r r r r r }
data & disk & $n/m$ & $n(I)$ & mem. & $t$ & search & m \\ 
\hline
wiki-nl\tnote{a} & & 1M/13M & 3--6 & 16 Gb & 16 & 7m & 8 \\
		&  &   & 3--6 & 5 Gb & 16 & 13m & 8  \\
		&  &   & 3--6 & 2 Gb & 1 & 25m & 8  \\
		&  &   & 10 & 11 Gb & 1 & 2h 41m & 0 \\
		& \checkmark  &  & 3--6 & 1 Gb & 1 &  1h 30m &  8 \\
\hline
wiki-en\tnote{b} & \checkmark & 12M/ 378M & 3--6 & 2 Gb & 1 & 6h 6m & 10 \\
 & \checkmark &  & 8 & 8 Gb & 1 & 6h 5m & 23 \\
 \hline
twitter\tnote{c} & \checkmark  & 53M/ 1963M & 3--6 & 6 Gb & 1 & 33h 19m & 0\\
 & \checkmark  &   & 7 & 8 Gb & 1 & 54h 26m & 0 \\
\hline
friendster\tnote{d} & \checkmark  & 68M/2586M & 3--6 & 6 Gb & 1 & 45h 2m & 68 \\ 
&\checkmark  & & 3--6 & 56 Gb & 9 & 8h 38m & 68 \\ 
&\checkmark  & & 10 & 7 Gb & 1 & 35h 7m & 57 \\ 
\hline
\end{tabular}
\begin{tablenotes}
\item[a]\cite{konect:2016:link-dynamic-nlwiki,konect:unlink}, multiple links were removed.
\item[b]\cite{konect:2016:dbpedia-link,konect:dbpedia2}, self-loops were removed.
\item[c]\cite{konect:2016:twitter,konect:twitter1}
\item[d]\cite{konect:2016:friendster}
\end{tablenotes}
\end{threeparttable}

\caption{The results of various runs of the algorithm on large data sets. The fifth column indicates the sizes of motifs that were sampled. The $t$ column shows the number of threads allowed to run concurrently. The last column indicates how many significant motifs were returned (under the EL model). The memory column indicates the maximum heapspace allowed for the Java Virtual Machine. Reported time does not include preloading. The $m$ column indicates the proportion of candidates that were motifs.}
\label{table:various}
\end{table}

\subsection{Large-Scale Motif Extraction}
\label{section:large}

\noindent Section~\ref{section:classification} showed that our method can, in principle, return characteristic motifs, even when used with the edgelist null-model. Since the codelength under the EL model can be computed very efficiently, this configuration should be extremely scalable. To test its limits, we run several experiments on large data sets ranging from a million to a billion links.  

In all experiments, we sample 1\hs000\hs000 motifs in total. We take the 100 most frequent motifs in this sample and compute their log-factors under the ER and EL models. We report the number of significant motifs found under the EL model. 

Table~\ref{table:various} shows the results. The largest data set that we can analyse stored in-memory with commodity hardware is the wiki-nl data set. For larger data sets, we store the graph on disk. Details are provided in the supplement.

This experiment shows we can perform motif analysis on data in the scale of billions of edges with very hardware. The sampling phase is `embarrassingly parallel', and indeed, a good speedup is achieved for multithreaded execution. We also observe that the amount of motifs found can vary wildly between data sets. The twitter and friendster data sets are from similar domains, and yet for twitter, no motifs are found, by a wide margin,\footnote{The EL model compressed better than the motif model by millions of bits in all cases.} whereas for friendster the majority of the subgraphs are motifs. What exactly causes the difference in these data sets is a matter of future research. 

With large data, using the full parallelism available is not always the best option. There is a trade-off between maximizing concurrency and avoiding garbage collection. The second line for the friendster data shows the fastest runtime (using the maximum stable heapspace) which used 9 concurrently running threads (with 16 logical cores available).

We also show that our method can scale to larger motifs, often with a modest increase in resources. This is highly dependent on the data, however. On the twitter data, sampling motifs larger than 7 did not finish within 48 hours. This may be due to an incomplete implementation of the Nauty algorithm: the data may contain subgraphs that take a long time to convert to their canonical ordering. A more efficient canonization algorithm (like the complete Nauty) could improve performance. However, as the results show, some data allows for fast analysis on larger motifs.

Preloading can be prohibitively expensive, in the same order as the analysis itself (this is not included in the reported runtimes). However, the graph in database format does not take up considerably more space than it does in raw edgelist-encoding. Preloading times could therefore be eliminated by distributing graph data in a suitable indexed binary format. 

\begin{figure}[tbh]
  \includegraphics[width=\linewidth]{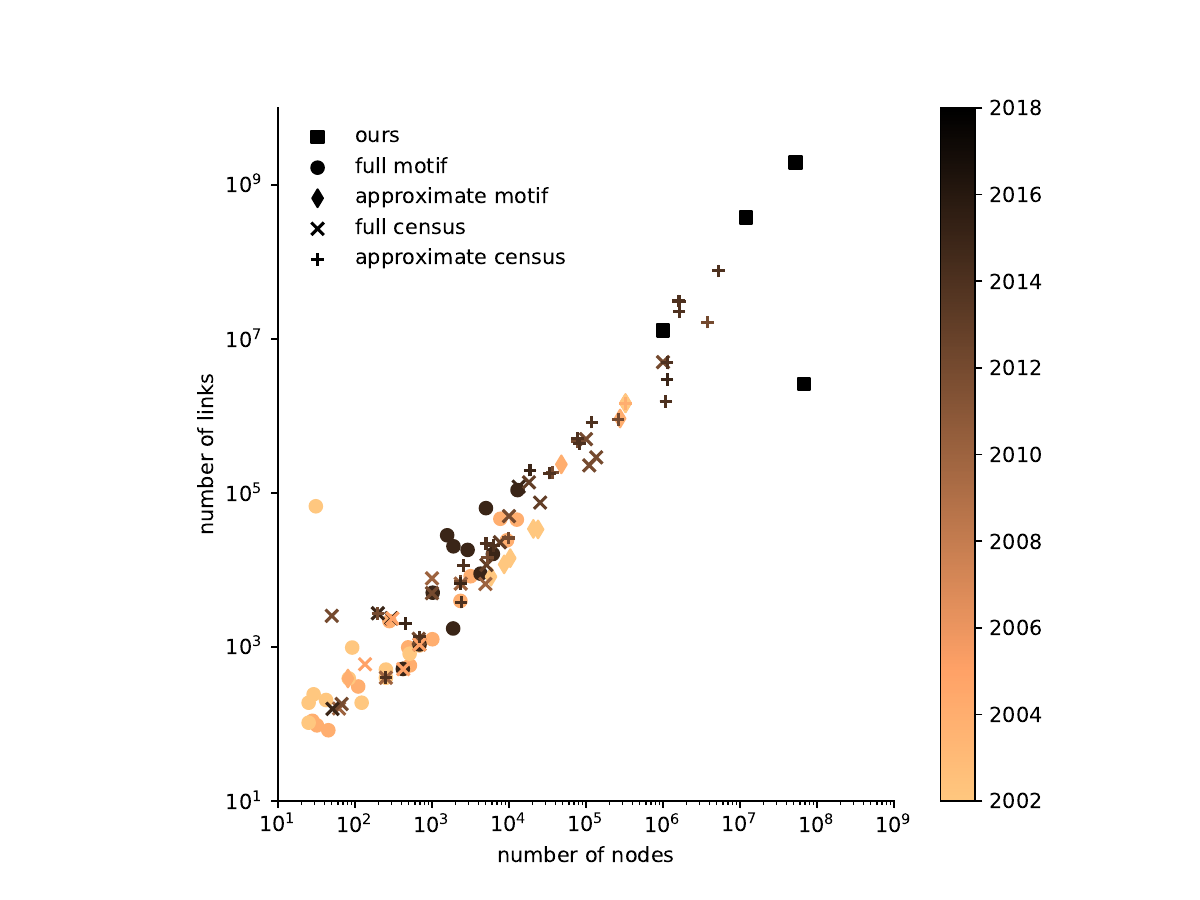}
  \caption{A scatterplot of previously published motif analyses. Note the logarithmic scale. Results are taken from references \cite{bhuiyan2012guise,hovcevar2014combinatorial,kashtan2004efficient,milo2004superfamilies,picard2008assessing,ribeiro2010g,carstens2013motifs,milo2002network,wang2014efficiently,li2012netmode,meira2014acc,paredes2015rand,slota2013fast,wang2012efficiently,wernicke2005faster}. \textbf{Full motif} analyses are those where the number of motifs is counted exactly on the data, and on a random ensemble. \textbf{Approximate motif} analyses are those where the count is approximated by sampling. A census performs the count, but no hypothesis test. We place our method somewhere between the full motif analysis and the approximate motif analysis: we make certain changes to the notion of a motif to achieve scalability, but our hypothesis test is fully correct (i.e. not an approximation). } 
\label{figure:others}
\end{figure}

For example, in the domain of knowledge graphs the HDT format by \cite{fernandez2013binary} provides both compression and indexing over the links of a graph.
Figure~\ref{figure:others} places these results in the context of currently published research.

\paragraph{Comparison to alternative methods} These results cannot be compared one-on-one to results from the literature: the protocol we follow differs in key places from the standard protocol envisioned for motif analysis. First, we do not check \emph{all} motifs of a given size, we use sampling frequency to make a preselection of candidates. Second, we focus on different statistics to determine what constitutes a motif. To still provide a broad sense of scale, we plot the sizes of graphs subjected to motif analysis in existing papers, together with our own in Figure~\ref{figure:others}. The collected data is available. \footnote{\url{https://github.com/pbloem/motive/blob/master/src/main/resources/data/motif-experiments.csv}} 

Note that there is a marked difference between the motif experiments and the census experiments. We suggest that this is not solely due to the extra cost of repeating the census on the random ensemble, but also due to the cost of just sampling the random ensemble. Such sampling is usually done through an MCMC method like the switching or the curveball algorithm. Such methods not only require a full copy of the data to be made for each sample, they also require a run-in of random transitions, until a proper mix is achieved. This mixing time increases with graph size, which means that even if the approximate census can be performed in constant time, producing the random ensemble becomes a bottleneck.

Of course, in our approach, we use the edgelist model instead of the degree-sequence model. If a null model can be found that similarly approximates the degree-sequence model, and allows for efficient sampling, the scale of approximate motif analysis may yet be extended.

\subsection{Scaling behavior}
\label{section:large}

\begin{figure}[tbh]
  \includegraphics[width=\linewidth]{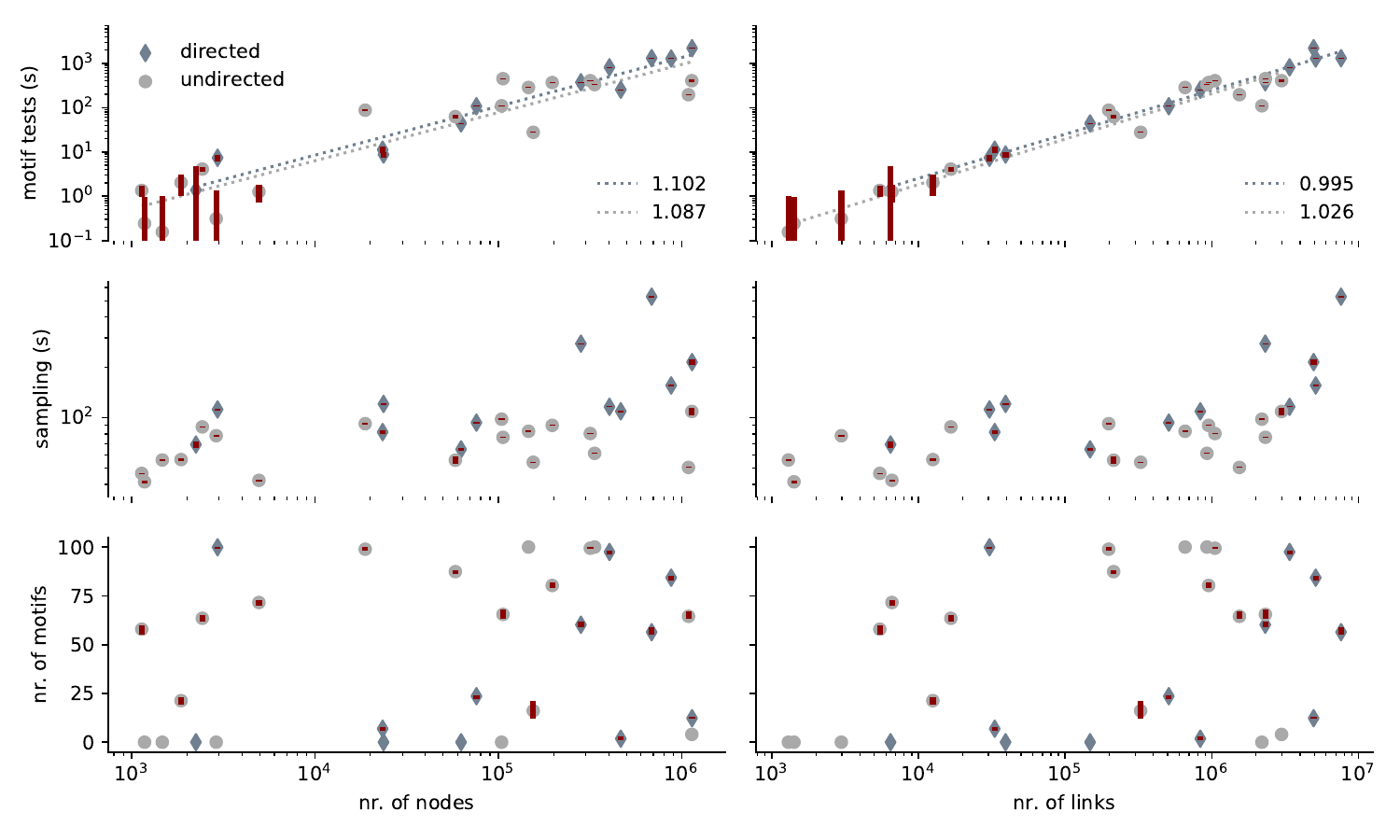}
  \caption{Top: Motif analysis runtimes for 30 graphs from the KONECT repository \cite{konect}.  Dotted lines show a linear fit in log space. The legend shows the slope. Middle: Sampling runtimes in the same experiment. Bottom: Number of motifs in the 100 candidate subgraphs. Error bars in the top and middle rows show the 95\% confidence interval over 10 repeats. For the bottom row, they show the \emph{range} of the data.} 
\label{figure:scaling}
\end{figure}

\noindent It is difficult to establish analytically how the method scales. To provide some insight, we  ran the full experiment on 30 medium-sized graphs from the KONECT repository \cite{konect}. For each, we sampled 1\hs000\hs000 instances, and performed a motif test on the top 100 candidates, using the EL model. We separate the runtime into the \emph{sampling} phase , and the \emph{motif testing phase}.

Figure~\ref{figure:scaling} shows the result. The pattern is noisy, but the motif testing phase admits a linear fit. We fitted a line to the logarithms of the values. The slope is close to one, suggesting that it is not unreasonable to expect linear scaling in both $n$ and $m$. This fits our expectation: the EL code is computed by a single pass over all nodes, to compute the sum in equation (\ref{eq:el-dir}) or (\ref{eq:el-undir}). This is usually the dominant routine in the algorithm.

The sampling, as expected, has a high variance between datasets, and low correlation with the size of the data. The four graphs for which sampling took the longest are large graphs, but these are also all web-graphs. It is not clear what causes this increase in runtime, since there are graphs of similar size for which sampling is fast.

\subsection{Conclusion} 

\label{section:conclusion}
We have presented a novel method for finding network motifs. Our method has several advantages:
\begin{itemize}
  \item The search for motif instances only needs to be run once: on the data $G$,  where the traditional approach requires a graph census to be repeated on samples from the null model.
 \item The search does not need to find all instances of a motif. We only require as many instances as can be found with the resources available. For large graphs, a relatively small number of instances may suffice to prove some motifs significant.
  \item This also allows us to retain a list of exactly those instances that made the subgraph a relevant motif. These can then be inspected by a domain-expert to establish whether the motif truly represents a ``functional unit''. 
  \item Given sufficiently strong evidence, a single test can be used to eliminate multiple null models. 
  \item The resulting relevance can be  used to compare the significance of motifs of different sizes in a meaningful way.
\end{itemize}
It is still a complicated question whether graph motifs, from this method or any other, represent a useful insight into the structure of the data. Our aim is to extend the method to knowledge graphs. Hopefully, in this setting, the resulting motifs will be easier to evaluate by domain experts.



\paragraph{Acknowledgements} We thank Pieter Adriaans for valuable discussions. This publication was supported by the Dutch national program COMMIT, by the Netherlands eScience center, and by the Amsterdam Academic Alliance Data Science (AAA-DS) Program Award to the UvA and VU Universities. 

%

\bibliographystyle{plain}
\bibliography{motifs}

\begin{thebibliography}{10}

\bibitem{aggarwal2014frequent}
Charu~C Aggarwal and Jiawei Han.
\newblock {\em Frequent pattern mining}.
\newblock Springer, 2014.

\bibitem{konect:dbpedia2}
Sören Auer, Christian Bizer, Georgi Kobilarov, Jens Lehmann, Richard Cyganiak,
  and Zachary Ives.
\newblock {DBpedia}: A nucleus for a web of open data.
\newblock In {\em Proc. Int. Semantic Web Conf.}, pages 722--735, 2008.

\bibitem{bhuiyan2012guise}
Mansurul~A Bhuiyan, Mahmudur Rahman, Mahmuda Rahman, and Mohammad Al~Hasan.
\newblock Guise: Uniform sampling of graphlets for large graph analysis.
\newblock In {\em 2012 IEEE 12th International Conference on Data Mining},
  pages 91--100. IEEE, 2012.

\bibitem{blitzstein2011sequential}
Joseph~K. Blitzstein and Persi Diaconis.
\newblock A sequential importance sampling algorithm for generating random
  graphs with prescribed degrees.
\newblock {\em Internet Mathematics}, 6(4):489--522, 2011.

\bibitem{bloem2018tutorial}
Peter Bloem and Steven de~Rooij.
\newblock A tutorial on mdl hypothesis testing for graph analysis.
\newblock {\em arXiv preprint arXiv:1810.13163}, 2018.

\bibitem{carstens2013motifs}
C.~J. Carstens.
\newblock Motifs in directed acyclic networks.
\newblock In {\em International Conference on Signal-Image Technology {\&}
  Internet-Based Systems, {SITIS} 2013, Kyoto, Japan, December 2-5, 2013},
  pages 605--611. {IEEE}, 2013.

\bibitem{carstens2016curveball}
Corrie~Jacobien Carstens, Annabell Berger, and Giovanni Strona.
\newblock Curveball: a new generation of sampling algorithms for graphs with
  fixed degree sequence.
\newblock {\em arXiv preprint arXiv:1609.05137}, 2016.

\bibitem{konect:coleman1957}
James Coleman, Elihu Katz, and Herbert Menzel.
\newblock The diffusion of an innovation among physicians.
\newblock {\em Sociometry}, pages 253--270, 1957.

\bibitem{cook1994substructure}
Diane~J. Cook and Lawrence~B. Holder.
\newblock Substructure discovery using minimum description length and
  background knowledge.
\newblock {\em CoRR}, cs.AI/9402102, 1994.

\bibitem{cover2006elements}
Thomas~M. Cover and Joy~A. Thomas.
\newblock {\em Elements of information theory {(2.} ed.)}.
\newblock Wiley, 2006.

\bibitem{fernandez2013binary}
Javier~D Fern{\'a}ndez, Miguel~A Mart{\'\i}nez-Prieto, Claudio Guti{\'e}rrez,
  Axel Polleres, and Mario Arias.
\newblock Binary rdf representation for publication and exchange (hdt).
\newblock {\em Web Semantics: Science, Services and Agents on the World Wide
  Web}, 19:22--41, 2013.

\bibitem{gehrke2003overview}
Johannes Gehrke, Paul Ginsparg, and Jon Kleinberg.
\newblock Overview of the 2003 kdd cup.
\newblock {\em ACM SIGKDD Explorations Newsletter}, 5(2):149--151, 2003.

\bibitem{charo2010efficient}
Charo I.~Del Genio, Hyunju Kim, Zolt{\'{a}}n Toroczkai, and Kevin~E. Bassler.
\newblock Efficient and exact sampling of simple graphs with given arbitrary
  degree sequence.
\newblock {\em CoRR}, abs/1002.2975, 2010.

\bibitem{gilbert1959random}
Edgar~N Gilbert.
\newblock Random graphs.
\newblock {\em The Annals of Mathematical Statistics}, pages 1141--1144, 1959.

\bibitem{grunwald2007minimum}
P.D. Gr{\"u}nwald.
\newblock {\em The minimum description length principle}.
\newblock The MIT Press, 2007.

\bibitem{hovcevar2014combinatorial}
Toma{\v{z}} Ho{\v{c}}evar and Janez Dem{\v{s}}ar.
\newblock A combinatorial approach to graphlet counting.
\newblock {\em Bioinformatics}, 30(4):559--565, 2014.

\bibitem{jha2015path}
Madhav Jha, C~Seshadhri, and Ali Pinar.
\newblock Path sampling: A fast and provable method for estimating 4-vertex
  subgraph counts.
\newblock In {\em Proceedings of the 24th International Conference on World
  Wide Web}, pages 495--505. ACM, 2015.

\bibitem{kashtan2004efficient}
Nadav Kashtan, Shalev Itzkovitz, Ron Milo, and Uri Alon.
\newblock Efficient sampling algorithm for estimating subgraph concentrations
  and detecting network motifs.
\newblock {\em Bioinformatics}, 20(11):1746--1758, 2004.

\bibitem{khakabimamaghani2013quatexelero}
Sahand Khakabimamaghani, Iman Sharafuddin, Norbert Dichter, Ina Koch, and Ali
  Masoudi-Nejad.
\newblock Quatexelero: an accelerated exact network motif detection algorithm.
\newblock {\em PloS one}, 8(7):e68073, 2013.

\bibitem{kiefer1953sequential}
Jack Kiefer.
\newblock Sequential minimax search for a maximum.
\newblock {\em Proceedings of the American Mathematical Society},
  4(3):502--506, 1953.

\bibitem{knerr1990single}
Stefan Knerr, L{\'e}on Personnaz, and G{\'e}rard Dreyfus.
\newblock Single-layer learning revisited: a stepwise procedure for building
  and training a neural network.
\newblock In {\em Neurocomputing}, pages 41--50. Springer, 1990.

\bibitem{konect:2014:moreno_names}
KONECT.
\newblock King james network dataset -- {KONECT}, October 2014.

\bibitem{konect:2015:moreno_innovation}
KONECT.
\newblock Physicians network dataset -- {KONECT}, April 2015.

\bibitem{konect:2016:friendster}
KONECT.
\newblock Friendster network dataset -- {KONECT}, October 2016.

\bibitem{konect:2016:twitter}
KONECT.
\newblock Twitter (www) network dataset -- {KONECT}, October 2016.

\bibitem{konect:2016:dbpedia-link}
KONECT.
\newblock Wikipedia, english network dataset -- {KONECT}, October 2016.

\bibitem{konect:2016:link-dynamic-nlwiki}
KONECT.
\newblock Wikipedia, nl (dynamic) network dataset -- {KONECT}, October 2016.

\bibitem{koskas2011nemo}
Michel Koskas, Gilles Grasseau, Etienne Birmel{\'e}, Sophie Schbath, and
  St{\'e}phane Robin.
\newblock Nemo: Fast count of network motifs.
\newblock {\em Book of Abstracts for Journ{\'e}es Ouvertes Biologie
  Informatique Math{\'e}matiques (JOBIM)}, pages 53--60, 2011.

\bibitem{konect}
J\'er\^ome Kunegis.
\newblock {KONECT} -- {The} {Koblenz} {Network} {Collection}.
\newblock In {\em Proc. Int. Conf. on World Wide Web Companion}, pages
  1343--1350, 2013.

\bibitem{konect:twitter1}
Haewoon Kwak, Changhyun Lee, Hosung Park, and Sue Moon.
\newblock What is {Twitter}, a social network or a news media?
\newblock In {\em Proc. Int. World Wide Web Conf.}, pages 591--600, 2010.

\bibitem{li2012netmode}
Xin Li, Douglas~S Stones, Haidong Wang, Hualiang Deng, Xiaoguang Liu, and Gang
  Wang.
\newblock Netmode: Network motif detection without nauty.
\newblock {\em PloS one}, 7(12):e50093, 2012.

\bibitem{mckay1981practical}
Brendan~D McKay et~al.
\newblock {\em Practical graph isomorphism}.
\newblock Department of Computer Science, Vanderbilt University Tennessee, US,
  1981.

\bibitem{meira2014acc}
Luis~AA Meira, Vin{\'\i}cius~R M{\'a}ximo, {\'A}lvaro~L Fazenda, and Arlindo~F
  Da~Concei{\c{c}}{\~a}o.
\newblock Acc-motif: accelerated network motif detection.
\newblock {\em IEEE/ACM Transactions on Computational Biology and
  Bioinformatics (TCBB)}, 11(5):853--862, 2014.

\bibitem{milo2004superfamilies}
Ron Milo, Shalev Itzkovitz, Nadav Kashtan, Reuven Levitt, Shai Shen-Orr, Inbal
  Ayzenshtat, Michal Sheffer, and Uri Alon.
\newblock Superfamilies of evolved and designed networks.
\newblock {\em Science}, 303(5663):1538--1542, 2004.

\bibitem{milo2002network}
Ron Milo, Shai Shen-Orr, Shalev Itzkovitz, Nadav Kashtan, Dmitri Chklovskii,
  and Uri Alon.
\newblock Network motifs: simple building blocks of complex networks.
\newblock {\em Science}, 298(5594):824--827, 2002.

\bibitem{newman2010networks}
Mark Newman.
\newblock {\em Networks: an introduction}.
\newblock Oxford University Press, 2010.

\bibitem{paredes2015rand}
Pedro Paredes and Pedro Ribeiro.
\newblock Rand-fase: fast approximate subgraph census.
\newblock {\em Social Network Analysis and Mining}, 5(1):1--18, 2015.

\bibitem{picard2008assessing}
Franck Picard, J-J Daudin, Michel Koskas, Sophie Schbath, and Stephane Robin.
\newblock Assessing the exceptionality of network motifs.
\newblock {\em Journal of Computational Biology}, 15(1):1--20, 2008.

\bibitem{konect:unlink}
Julia Preusse, J\'er\^ome Kunegis, Matthias Thimm, Thomas Gottron, and Steffen
  Staab.
\newblock Structural dynamics of knowledge networks.
\newblock In {\em Proc. Int. Conf. on Weblogs and Social Media}, 2013.

\bibitem{reguly2006comprehensive}
Teresa Reguly, Ashton Breitkreutz, Lorrie Boucher, Bobby-Joe Breitkreutz,
  Gary~C Hon, Chad~L Myers, Ainslie Parsons, Helena Friesen, Rose Oughtred, Amy
  Tong, et~al.
\newblock Comprehensive curation and analysis of global interaction networks in
  saccharomyces cerevisiae.
\newblock {\em Journal of biology}, 5(4):11, 2006.

\bibitem{renyi1959random}
A~Renyi and P~Erd\H{o}s.
\newblock On random graphs.
\newblock {\em Publicationes Mathematicae}, 6(290-297):5, 1959.

\bibitem{ribeiro2010g}
Pedro Ribeiro and Fernando Silva.
\newblock G-tries: an efficient data structure for discovering network motifs.
\newblock In {\em Proceedings of the 2010 ACM Symposium on Applied Computing},
  pages 1559--1566. ACM, 2010.

\bibitem{rissanen1978modeling}
Jorma Rissanen.
\newblock Modeling by shortest data description.
\newblock {\em Automatica}, 14(5):465--471, 1978.

\bibitem{rissanen1979arithmetic}
Jorma Rissanen and Glen~G Langdon.
\newblock Arithmetic coding.
\newblock {\em IBM Journal of research and development}, 23(2):149--162, 1979.

\bibitem{ristoski2016collection}
Petar Ristoski, Gerben Klaas~Dirk de~Vries, and Heiko Paulheim.
\newblock A collection of benchmark datasets for systematic evaluations of
  machine learning on the semantic web.
\newblock In {\em International Semantic Web Conference}, pages 186--194.
  Springer, 2016.

\bibitem{konect:harrison}
Christoph R\"{o}mhild and Chris Harrison.
\newblock \url{http://chrisharrison.net/projects/bibleviz/index.html}, 2007.
\newblock Accessed: 2014-08-22.

\bibitem{slota2013fast}
George~M Slota and Kamesh Madduri.
\newblock Fast approximate subgraph counting and enumeration.
\newblock In {\em Parallel Processing (ICPP), 2013 42nd International
  Conference on}, pages 210--219. IEEE, 2013.

\bibitem{slota2014complex}
George~M Slota and Kamesh Madduri.
\newblock Complex network analysis using parallel approximate motif counting.
\newblock In {\em Parallel and Distributed Processing Symposium, 2014 IEEE 28th
  International}, pages 405--414. IEEE, 2014.

\bibitem{strona2014fast}
Giovanni Strona, Domenico Nappo, Francesco Boccacci, Simone Fattorini, and
  Jesus San-Miguel-Ayanz.
\newblock A fast and unbiased procedure to randomize ecological binary matrices
  with fixed row and column totals.
\newblock {\em Nature communications}, 5, 2014.

\bibitem{leeuwen2006compression}
Matthijs van Leeuwen, Jilles Vreeken, and Arno Siebes.
\newblock Compression picks item sets that matter.
\newblock In Johannes F{\"{u}}rnkranz, Tobias Scheffer, and Myra Spiliopoulou,
  editors, {\em Knowledge Discovery in Databases: {PKDD} 2006, 10th European
  Conference on Principles and Practice of Knowledge Discovery in Databases,
  Berlin, Germany, September 18-22, 2006, Proceedings}, volume 4213 of {\em
  Lecture Notes in Computer Science}, pages 585--592. Springer, 2006.

\bibitem{wang2014efficiently}
Pinghui Wang, John Lui, Bruno Ribeiro, Don Towsley, Junzhou Zhao, and Xiaohong
  Guan.
\newblock Efficiently estimating motif statistics of large networks.
\newblock {\em ACM Transactions on Knowledge Discovery from Data (TKDD)},
  9(2):8, 2014.

\bibitem{wang2012efficiently}
Yuyi Wang and Jan Ramon.
\newblock An efficiently computable support measure for frequent subgraph
  pattern mining.
\newblock {\em Machine Learning and Knowledge Discovery in Databases}, pages
  362--377, 2012.

\bibitem{wernicke2005faster}
Sebastian Wernicke.
\newblock A faster algorithm for detecting network motifs.
\newblock In Rita Casadio and Gene Myers, editors, {\em Algorithms in
  Bioinformatics, 5th International Workshop, {WABI} 2005, Mallorca, Spain,
  October 3-6, 2005, Proceedings}, volume 3692 of {\em Lecture Notes in
  Computer Science}, pages 165--177. Springer, 2005.

\end{thebibliography}

\pagebreak
\appendix
\section{Supplement}

\subsection{Hardware}

All experiments were run on a single machine with a 2.60 Ghz Intel Xeon processor (E5-2650 v2) with 64 Gigabytes of memory and 8 physical cores. The memory and cores available to the program differ per experiment and are reported where relevant. 

\subsection{Integer and sequence codes}

\paragraph{Encoding integers and sequences} In the following, we will often need to encode single natural numbers, or a sequence of natural number from a finite range. For single numbers, we will use the code corresponding to the probability distribution $p^\N(n) = 1/ (n(n+1))$, and denote it $L^\N(n)$.


For sequences of elements from a finite set, we use the code corresponding to a \emph{Dirichlet-Multinomial} (DM) distribution. Let $S$ be a sequence of length $k$ of elements from some alphabet $\Sigma$. Conceptually, the DM distribution models the following sampling process: we sample a probability vector $p$ on $[0, |\Sigma|]$ from a Dirichlet distribution with parameter vector $\alpha$, and then sample $k$ symbols from the categorical distribution represented by $p$. The probability mass function corresponding to this process can be expressed as
\begin{align*}
&p^\text{DirM}_\alpha(S\mid k, \Sigma) = \prod_{i \in [1,k]} \text{DirM}_\alpha(S_i\mid S_{1:i-1}, k, \Sigma) \\
&\text{DirM}_\alpha(S_i \mid S', k, \Sigma) = \frac{f(S_i, S') + \alpha_i}{|S'| + \sum_i \alpha_i}
\end{align*}
where $f(x, X)$ denotes the frequency of $x$ in $X$. We use $\alpha_i = 1/2$ for all $i$. Let $L^\text{DirM}_{k,\Sigma} (S) = -\log p^\text{DirM}(S \mid k, \Sigma)$. The DM model can be seen as encoding each element from $S_i$, using the smoothed relative frequency of $S_i$ in the subsequence $S_{1:i=1}$ preceding it. Thus the probability of a given symbol changes at each point in the sequence, based on how often it has been observed up to that point.

Note that this code is parametrized with $k$ and $\Sigma$. If these cannot be deduced from information already stored, they need to be encoded separately. When encoding natural numbers, we will have $\Sigma = [0, n_\text{max}]$, and we only need to encode $n_\text{max}$. A useful property of the DM code is that it is \emph{exchangeable}: if we re-arrange the elements of $S$, the codelength remains the same. 

Note that, since we use $L^\N(n)$ and $L^\text{DirM}(n)$ only in the motif code, there is no need for them to be optimal. The better they compress, the more motifs we will find, but we do not require optimal results for the algorithm to be valid.

\subsection{Sampling algorithm}

For the first experiment, we use the following algorithm to sample a graph with $k$ injected motifs.

We use the following procedure to sample an undirected graph with $5\,000$ nodes and $10\,000$ links, containing $k$ injected instances of a particular motif $M$ with $n'$ nodes and $m'$ links. Let $M$ be given (in our experiment $M$ is always the graph indicated in red in Figure~\ref{figure:plot-synthetic}).

\begin{enumerate}
  \item Let $n = 5000 - (n'-1)k$ and $m = 10000 - m'k$ and sample a graph $H$ from the uniform distribution over all graphs with $n$ nodes and $m$ links.   
  \item Label $k$ random nodes, with degree 5 or less, as instance nodes.
  \item Let $p^\text{cat}$ be a categorical distribution on $\{1, \ldots, 5\}$, chosen randomly from the uniform distribution over all such distributions.
  \item Label every connection between an instance node and a link with a random value from $p^\text{cat}$. Links incident to two instance nodes, will thus get \emph{two} values.
  \item Reconstruct the graph $G$ from $M$ and $H$.
\end{enumerate}

\subsection{Datasets}
In the second experiment, the following datasets are used.

\begin{description}
\item[kingjames (undirected, $n=1773, m=9131$)] Co-occurrences of nouns in the text of the King James Bible \cite{konect:2014:moreno_names,konect:harrison}.
\item[yeast (undirected, $n=1528, m=2844$)] A network of the protein interactions in yeast, based on a literature review \cite{reguly2006comprehensive}. 
\item[physicians (directed, $n=241, m=1098$)] Nodes are physicians in Illinois \cite{konect:2015:moreno_innovation,konect:coleman1957}. 
\item[citations (directed, $n=1769, m=4222$)] The arXiv citation network in the category of theoretical astrophysics, as created for the 2003 KDD Cup \cite{gehrke2003overview}. \footnotemark \end{description}
All data sets are simple (no multiple edges, no self-loops). In each case we take $5 \cdot 10^6$ samples with $n_\text{min} = 3$ and $n_\text{max} = 6$. We test the 100 motifs with the highest number of instances (after overlap removal), and report the log-factor for each null model. For the edgelist and ER models we use a Fibonacci search at full depth, for the degree-sequence model we restrict the search depth to $3$. For the degree-sequence estimator, we use $40$ samples and $\alpha=0.05$ to determine our confidence interval. We use the same set of instances for each null model.

\footnotetext{We follow the procedure outlined in \cite{carstens2013motifs}: we include only papers before 1994, remove forward citations, and select the largest connected component. 
}

\subsection{Experimental details of disk-based experiments}

The following details pertain to running the motif analysis using a disk-based graph-store.

The graph is stored in two lists, as it is in the in-memory version. The first, the \emph{forward list}, contains at index $i$ a sorted list of integers $j$ for all links $(n_i,n_j)$ that exist: i.e. a list of outgoing neighbors of $n_i$. The second, the \emph{backward list}, contains lists of incoming neighbors for each node. The data is stored on disk in a way that allows efficient random access (using the MapDB database engine\footnotemark).

\footnotetext{\url{http://www.mapdb.org/}}

For large graphs, converting a file from the common edgelist encoding (a file with a line for each link, encoded as a pair of integers) to this format can take considerable time, but this needs to be done only once, so we show the preloading and analysis times separately. Loading the graph is done by performing a disk-based sort of the edgelist-encoded file, on the first element of each pair, loading the forward list, sorting again by the second element, and loading the backward list. This minimizes random access as both lists can be filled sequentially in one pass.

We only require one pass over the whole data, to compute the model parameters (eg. the degree sequence). For the samplixng and the computation of the log factors only relatively small amounts of random access are required. Since a graph can, in principle, be compressed with only a very small number of instances of a given motif, this gives us a very scalable method to find motifs in large data. 

For disk-based experiments, we limit the total number of rewritten links in the template graph to 500 000, to limit memory use. If the motif with a given list of instances results in more rewritten links, we do not consider it. Note that, since we search for a good pruning of the instance list, the motif will still be considered with a more heavily pruned instance list. 
A large number of rewritten links suggest that there are many instances with high ex-degree, so we likely do not lose much by this heuristic. 

\end{document}